\documentclass[runningheads]{llncs}
\usepackage{svg}
\usepackage{graphicx}
\usepackage[export]{adjustbox}
\usepackage{enumerate}
\usepackage{amssymb}
\usepackage{floatrow}
\usepackage{amsmath}
\usepackage{multirow}
\usepackage{booktabs}
\usepackage{textgreek}

\usepackage{hhline}
\usepackage{array}
\usepackage{booktabs} 
\usepackage{tabularx} 
\usepackage{caption}  
\usepackage{floatrow} 
\usepackage{graphicx,subfigure}
\usepackage{tabu}
\usepackage{siunitx}

\usepackage{verbatim} 
\usepackage{caption} 
\usepackage{color}
\usepackage{url}
\usepackage[colorlinks = true]{hyperref}
\begin{document}
%
%
\title{\texttt{SketchGPT}: Autoregressive Modeling for Sketch Generation and Recognition}
\titlerunning{Autoregressive Modeling for Sketch Generation and Recognition}
%
\author{Adarsh Tiwari\inst{1} \inst{2}\orcidID{0009-0004-9266-7175}\and
Sanket Biswas\inst{1} \inst{2}\thanks{Main Corresponding Author}\orcidID{0000-0001-6648-8270} \and
Josep Lladós\inst{1} \inst{2}\orcidID{0000-0002-4533-4739} 
}

\authorrunning{A.Tiwari et al.}
\institute{Computer Vision Center, Catalonia, Spain \\
               \email{adarshtd17@gmail.com, \{sbiswas, josep\}@cvc.uab.es} 
               \and
               Computer Science Department \\ 
               Universitat Autònoma de Barcelona, Catalonia, Spain }
\maketitle              
\begin{abstract}

We present \textbf{SketchGPT}, a flexible framework that employs a sequence-to-sequence autoregressive model for sketch generation, and completion, and an interpretation case study for sketch recognition. By mapping complex sketches into simplified sequences of abstract primitives, our approach significantly streamlines the input for autoregressive modeling. SketchGPT leverages the next token prediction objective strategy to understand sketch patterns, facilitating the creation and completion of drawings and also categorizing them accurately. This proposed sketch representation strategy aids in overcoming existing challenges of autoregressive modeling for continuous stroke data, enabling smoother model training and competitive performance. Our findings exhibit SketchGPT's capability to generate a diverse variety of drawings by adding both qualitative and quantitative comparisons with existing state-of-the-art, along with a comprehensive human evaluation study. The code and pretrained models will be released on this GitHub$^\dagger$. 

\def\thefootnote{$\dagger$}\footnotetext{{\url{https://github.com/eltonjohnfanboy/SketchGPT}}}

\keywords{Sketch Completion  \and Sketch Recognition \and Next Stroke Prediction \and Stroke-to-Primitive Mapping \and Autoregressive Models}
\end{abstract}
\section{Introduction}\label{s:intro}
Sketches serve as foundational conceptual elements drawn by humans for a broad spectrum of applications~\cite{jonson2005design,suwa2022roles}, ranging from technical diagrams in architecture~\cite{de2015cvc}, electronics~\cite{rusinol2010relational}, mechanical engineering~\cite{egiazarian2020deep}, to educational tools and even in entertainment, as seen in the popular games like Pictionary~\cite{bhunia2020pixelor}. In the Graphics Recognition domain, the interpretation of hand drawn symbols and sketches has gained special attention. 

With the growth of Large Language Models (LLMs)~\cite{brown2020language,touvron2023llama} for the interpretation and generation of meaningful textual documents~\cite{lv2023kosmos}, it has also opened doors to a wide array of parallel applications. Among these applications is the understanding of sketches, which can be compared to interpreting sentences generated by a unique grammar or a visual language model~\cite{alayrac2022flamingo}. This draws an interesting parallel, just as the interpretation of a handwritten text involves recognizing letters or characters as basic building blocks, \textit{understanding sketches demand recognition of the composed objects that can be associated to bi-dimensional lexical units or strokes}. Unlike static images, sketches are recorded as a sequence of dynamic pen tip movements which follows a specific order and structure on the canvas, and then finally post-processed and rendered in the image space. Exploring this sequential nature of sketches, similar to a language model, can help to reveal the hidden lexicon encoded in every stroke. Previous attempts like SketchRNN~\cite{Ha2018} showcased the importance of sequential stroke information for sketch generation, while Sketch-BERT~\cite{lin2020sketch} extended this approach to sketch recognition and retrieval by adapting BERT's language modeling~\cite{devlin2018bert} techniques. 

Despite these contributions, the field still requires a holistic model capable of addressing the spectrum of tasks associated with sketches, from generation to classification and beyond, without the need to rely on a specialized encoder for bidirectional context understanding. This work introduces \textit{SketchGPT}, an autoregressive generative model inspired by the emergent capabilities of the GPT (Generative Pretrained Transformer) model series~\cite{brown2020language,radford2018improving,radford2019language} in the domain of text. The key advantages of the next element prediction pretraining principle over existing sketch language modeling methods~\cite{Ha2018,lin2020sketch} are: 1) Such principle is highly suitable for sketch generation, where \textit{the order and sequence of strokes are essential}. Each stroke in a sketch depends on the preceding strokes, which makes it a strong candidate. 2) Sketches can vary greatly in complexity and style, from simple line drawings to more complex object compositions. Autoregressive models can thrive in handling this variability, as they learn to \textit{predict a wide range of possible next steps in the sequence}, making them versatile for modeling different types of sketches. 3) These models can adjust the generation process at each step based on the previously generated content, to \textit{offer a higher level of precision for capturing the fine-grained information} of sketches. 

However, autoregressive approaches can sometimes negatively impact the reconstruction and prediction scores as they are prone to overfitting to the actual data through teacher forcing, and may also struggle to accurately complete drawings when they have to rely on their own generated predictions~\cite{aksan2020cose}. Hence, to make the model generalizable, we propose a \textit{stroke-to-primitive mapping} strategy inspired by ~\cite{alaniz2022abstracting} which allows discretizing sketches into a finite set of abstract primitives, simplifying the learning process and reducing overfitting. This approach ensures more consistent and structured predictions, minimizing errors in continuing drawings from inaccurate or incomplete predictions. 

Our main contributions are: 1) We present a GPT-inspired autoregressive model that learns neural representations of sketches, capturing their dynamic drawing process. 2) We propose a stroke-to-primitive abstraction strategy to simplify input data and enhance model generalization across diverse sketches. 3) We present a multi-task model capable of predicting the next stroke, generating, completing, and recognizing sketches, showcasing its overall versatility in sketch-related tasks. 4) Empirically, we also propose a quantitative study for sketch generation to compare with state-of-the-art models, and a comprehensive human evaluation study to assess the quality of generated sketches.

\section{Related Work}\label{s:soa}

\noindent
\textbf{Sketch as a Language:} Just as language comprises a set of syntactic rules and semantic structures, sketches can be interpreted as visual sentences~\cite{ganin2021computer,mas2010syntactic}, constructed through a series of strokes or "words" that follow a certain grammar. This analogy underpins our approach to modeling sketches, where the sequential nature of drawing parallels the linear construction of sentences in natural language. SketchRNN~\cite{Ha2018} employed the first autoregressive model based on LSTM~\cite{hochreiter1997long} to understand and generate sketches based on their sequential stroke data. Following this, SketchBERT~\cite{lin2020sketch} leveraged the power of BERT's language modeling objective for tasks such as sketch recognition and retrieval, showcasing the applicability of NLP techniques in the sketch domain. Recently, SketchKnitter~\cite{wang2022sketchknitter} proposed a diffusion model to understand real human sketch distributions, formulating sketch generation as a denoising process where the generator adjusts random stroke points over time to create a recognizable sketch. Our work builds upon learning neural representations of vectorized sketches in a GPT-like autoregressive way with next stroke prediction task, so it can be extended to multiple downstream tasks like sketch recognition, completion and generation. 

\noindent
\textbf{Sketch Abstraction with Primitives:} Simplifying complex shapes into basic primitives is a key strategy for understanding the environment~\cite{biederman1987recognition} mimicking humans, a concept used in various fields like CAD sketches~\cite{ganin2021computer}, vector-like bitmap images~\cite{reddy2021im2vec}and 3D shape reconstruction from both sketches~\cite{smirnov2019deep} and images~\cite{liu2018physical}. Previous attempts for abstract sketch generation~\cite{muhammad2019goal,Muhammad_2018_CVPR} developed reinforcement learning paradigm of a stroke removal policy that learns to predict which strokes can be safely removed without affecting recognizability. While recently Alaniz \textit{et. al.}~\cite{alaniz2022abstracting} proposed a Primitive Matching Network which learns to map each stroke of a sketch to its most similar primitive in a given set with a distance transform loss, predicting an affine transformation which can align the selected primitive to the target stroke. Inspired by~\cite{alaniz2022abstracting} we propose a simple sketch-to-primitive mapping strategy that helps us to simplify and discretize the representation space of the input sketch, and improve our autoregressive training strategy. 

\noindent
\textbf{Sketch Applications:} The introduction of the TU-Berlin~\cite{eitz2010sketch}, Sketchy~\cite{sangkloy2016sketchy} and QuickDraw~\cite{jongejan2016quick} datasets has significantly propelled research in sketch recognition and generation. Initially, the challenge was tackled using maximal margin classifiers fused with hand crafted features~\cite{schneider2014sketch,li2013sketch}. However, the introduction of large-scale sketch datasets~\cite{jongejan2016quick,sangkloy2016sketchy} paved the way for deep learning models~\cite{song2017deep,yu2016sketch}, which has achieved performance surpassing that of humans. Other application domains of interest include sketch-based image retrieval~\cite{dey2019doodle}, creative sketch generation~\cite{bhunia2022doodleformer}, sketch-based object localization~\cite{tripathi2020sketch} where abstract sketches serve as conceptual representations. Employing transformer-based approaches~\cite{bhunia2022doodleformer,lin2020sketch,ribeiro2020sketchformer,tiwari2023can} have been the current state-of-the-art practice, and innovating with learnt tokenization to enhance sketch interpretability.

\section{Method}\label{s:method}



\begin{figure}[t]
    \centering
    \includegraphics[width=0.8\linewidth, fbox]{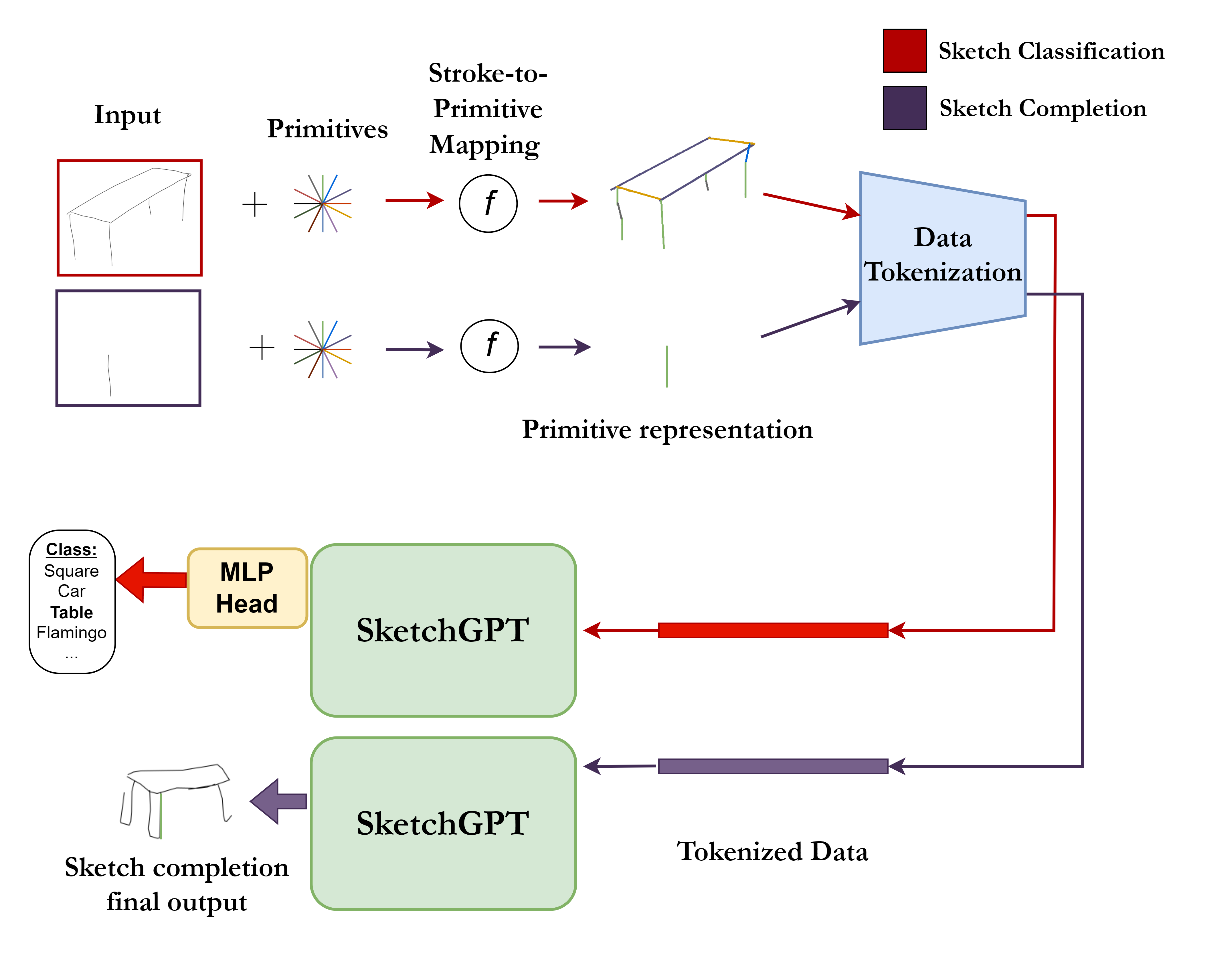}
    \caption{\textbf{Overview of \textit{SketchGPT}.} Given an input sketch (complete or incomplete), the model predicts the next strokes for incomplete sketches or classifies the complete ones, effectively adapting to different tasks.}
    \label{fig:fw_overview}
\end{figure}

\noindent
Built upon the foundations of autoregressive GPT-like models, \textit{SketchGPT} is a task-agnostic generative transformer pre-trained on the \textit{QuickDraw} dataset~\cite{Ha2018} containing multiple object categories of sketches. The model learns neural representations of sketch data, capturing sequential dependencies among their compounding strokes. The pre-trained representations learned on multiple concepts exhibit adaptability to a wide range of downstream sketch application tasks. Our focus in this work is aimed mainly towards solving \textit{sketch completion, sketch generation} and \textit{sketch classification} tasks. In Fig.~\ref{fig:fw_overview}, we show an overview of the complete framework, focusing on the aforementioned downstream tasks. Sketch generation and completion can be seen as two instances of the same model, thus for the sake of simplicity, we show a single model in this figure. For both of these tasks, namely classification and completion, we initiate the process with a sketch abstraction step that encodes the sketch strokes, in stroke-3 format, as sequences of primitives. A predefined finite alphabet of straight-line-based tokens has been used as primitives. The discretization of strokes reduces the complexity of the model in its learning phase and additionally produces tokenization compatible with the requirements of the language-based model. The tokenization step yields a feature vector that serves as an input feed for the fine-tuned models to conduct the downstream task. In subsequent subsections, detailed explanations are provided for the aforementioned steps.


\subsection{Data Preprocessing and Sketch Abstraction}

At the basic level, a sketch is a collection of time-stamped coordinate points obtained by a digital pen device. Some representation models compactly reduce this sequence of coordinates to polygonal approximations utilizing a vectorization process. Let us refer as {\em strokes} the segments of this approximation. The {\tt QuickDraw} dataset \cite{Ha2018}, used in this work as the benchmark, represents sketches by the \textit{stroke-3} encoding. In this format, each sketch is stored as a list of points, corresponding to the strokes, i.e. the segments of the vectorial approximation. The points consist of three values: ($x$,$y$,$p$). The elements $x$ and $y$ are the coordinate offsets, and $p$ represents the status of the pen ({\em on paper} or {\em on air}). To ensure uniformity in the data we perform min-max normalization on the coordinate values, restricting them to the normalized range of 0 to 1 (\(x, y \in [0, 1]\)).

\begin{figure}[t]
    \centering
    \includegraphics[width=\linewidth]{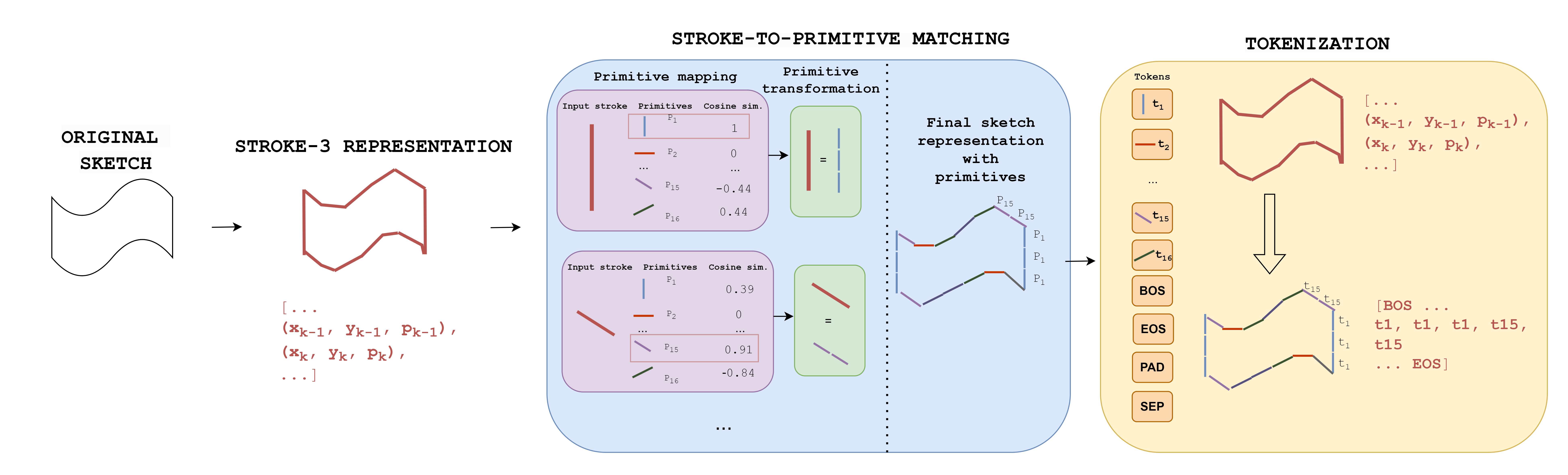}
    \caption{\textbf{Illustration of the stroke-to-primitive mapping and tokenization process.} We observe how raw stroke data is converted into a more structured representation, involving the interpretation of stroke primitives and then further tokenizing to be fed to the SketchGPT framework. }
    \label{fig:primitives}
\end{figure}

The sketch abstraction phase, inspired by the work of Alaniz \textit{et al.}\cite{alaniz2022abstracting} conducts a stroke-to-primitive mapping process that discretizes the strokes of the normalized sketch with a finite set of predefined primitives (Fig. \ref{fig:primitives}). In stroke-3 data, a stroke is a straight line defined by the pair of the coordinates of subsequent tuples \((x_i, y_i)\) and \((x_{i+1}, y_{i+1})\). Since the proposed architecture is inspired by a GPT-based language model, we are required to encode the sketch using a discrete set of tokens. With this aim, each stroke has been approximated by a primitive out of a dictionary. The dictionary contains a list of predefined primitives at regular orientation shifts and fixed length.

Each stroke \(s_i\) is mapped to its most similar primitive \(p_j\), in terms of the cosine similarity between the corresponding orientations, computed as follows:

\begin{equation}
\text{sim}(s_i, p_j) = \frac{s_i \cdot p_j}{\|s_i\| \cdot \|p_j\|} 
\label{equ:sim}
\end{equation}

where \(s_i \cdot p_j\) is the dot product of the vectors \(s_i\) and \(p_j\), and \(\|s_i\|\) and \(\|p_j\|\) denote the Euclidean norms of the vectors \(s_i\) and \(p_j\), respectively. The primitive \(p_i\) assigned to the stroke \(s_i\) is the one that maximizes this similarity measure:

\begin{equation}
p_i = \arg \max_{p_j \in P} \text{sim}(s_i, p_j) 
\label{eq:max}
\end{equation}



The primitives of the dictionary have a fixed length. Since the strokes of the input sketch are generated in a vectorization process, their lenght is variable, depending on the smoothness of the original curve. To better approximate the final encoding, we introduce a scaling transformation \(T\), that adjusts the length of the selected primitive \(p_i\) with the length of the original stroke \(s_i\):
\begin{equation}
T(p_i, s_i) = \begin{pmatrix} \left\lceil \frac{m(s_i)}{m(p_i)} \right\rceil \end{pmatrix}    
\label{eq:scaling}
\end{equation}

where \(m(s_i)\) represents the length of the input stroke, and the ceiling operation (\(\left\lceil \frac{m(s_i)}{m(p_i)} \right\rceil\)) guarantees that the length of the primitive \(p_i\) aligns to the nearest integer length of the original stroke \(s_i\).
After the entire process, the resulting representation for a sketch \(S_i\) can be expressed as:

\begin{equation}
S_i = \{ p_1 \cdot T(p_1, s_1), p_2 \cdot T(p_2, s_2), \ldots, p_n \cdot T(p_n, s_n) \} 
\end{equation}

We also make use of this information to tokenize the data. Our tokenization method establishes a vocabulary \(V\) that consists of tokens representing the primitives (\(p_i\)), enriched with four special tokens: \( \text{BOS} \) (beginning of sequence), \( \text{SEP} \) (separator), \( \text{EOS} \) (end of sequence), and \( \text{PAD} \) (padding). Let \( \tau \) denote the vector used for tokenization, constructed by repeating each primitive \( p_i \) according to its scaling factor:
\begin{equation}
\mathcal{T} = [BOS, p_1, \ldots, p_1, p_2, SEP, p_2, \ldots, p_2, \ldots, SEP, p_n, \ldots, p_n, EOS]
\end{equation}

where the repetition of each primitive \( p_i \) is determined by its scaling factor \(T(p_i, s_i)\) defined in Eq. \ref{eq:scaling}. This methodology ensures that the tokenization incorporates the crucial details of our sketch, creating a representation that can be fed into the model.

\subsection{Model Architecture} 

\textit{SketchGPT} is an autoregressive decoder-only based transformer model inspired in the GPT-2 architecture design. The fundamental element of such architectures, as well as in our proposed model, is the \textit{causal masked multi-head self-attention mechanism}. Masked self-attention is a specialized adaptation of the conventional self-attention mechanism. It adds a constraint through which each element (a stroke) of our input sequence can only attend to the previous ones, which enables them to get trained autoregressively, where the prediction of the next token relies exclusively on the preceding context. Given a sequence \(X\) which represents a sketch, the following equation models this masking mechanism:

\begin{equation}
\text{MaskedAttention}(X) = \text{softmax}\left(\frac{(X \cdot W_Q \cdot (X \cdot W_K)^T) \odot \text{Mask}}{\sqrt{d_k}}\right) \cdot (X \cdot W_V)
\end{equation}




where \(W_Q\), \(W_K\), and \(W_V\) are learnable weight matrices, and \(d_k\) is the dimension of the query and key vectors. The operation involves computing query (\(Q = X \cdot W_Q\)), key (\(K = X \cdot W_K\)), and value (\(V = X \cdot W_V\)) projections of the input sequence, followed by a scaled dot-product attention and a weighted sum using the softmax function. The \(\odot\) denotes element-wise multiplication, and \(\text{Mask}\) is a binary mask matrix that prevents attending to future positions, preserving the temporal order of the sequence.

Given a sequence \(X\), the multi-head self-attention operation computes multiple parallel self-attention blocks (or heads) as shown below:

\begin{equation}
\text{MultiHead}(X) = \text{Concat}(\text{head}_1, \text{head}_2, \ldots, \text{head}_h) \cdot W_O
\end{equation}

where \(\text{head}_i = \text{MaskedAttention}_i(X)\) represents the causal masked self-attention output for the \(i\)-th head. The resulting concatenated representation is linearly transformed by the output projection matrix \(W_O\). This enrichment in the architecture through the use of multiple heads boosts the model's efficacy in capturing diverse contextual and temporal relationships within the sequence. After the multi-head attention layer, the output is directed through a Multi-Layer Perceptron (MLP), giving place to the fundamental building block of the network, which is the transformer block. Stacking these transformer blocks shapes the core of the transformer block used in our work.

\subsection{Pre-training method for SketchGPT}

\begin{figure}[t]
\centering     
\subfigure[Pre-Training for Completion]{\label{fig:ft}\includegraphics[width=60mm]{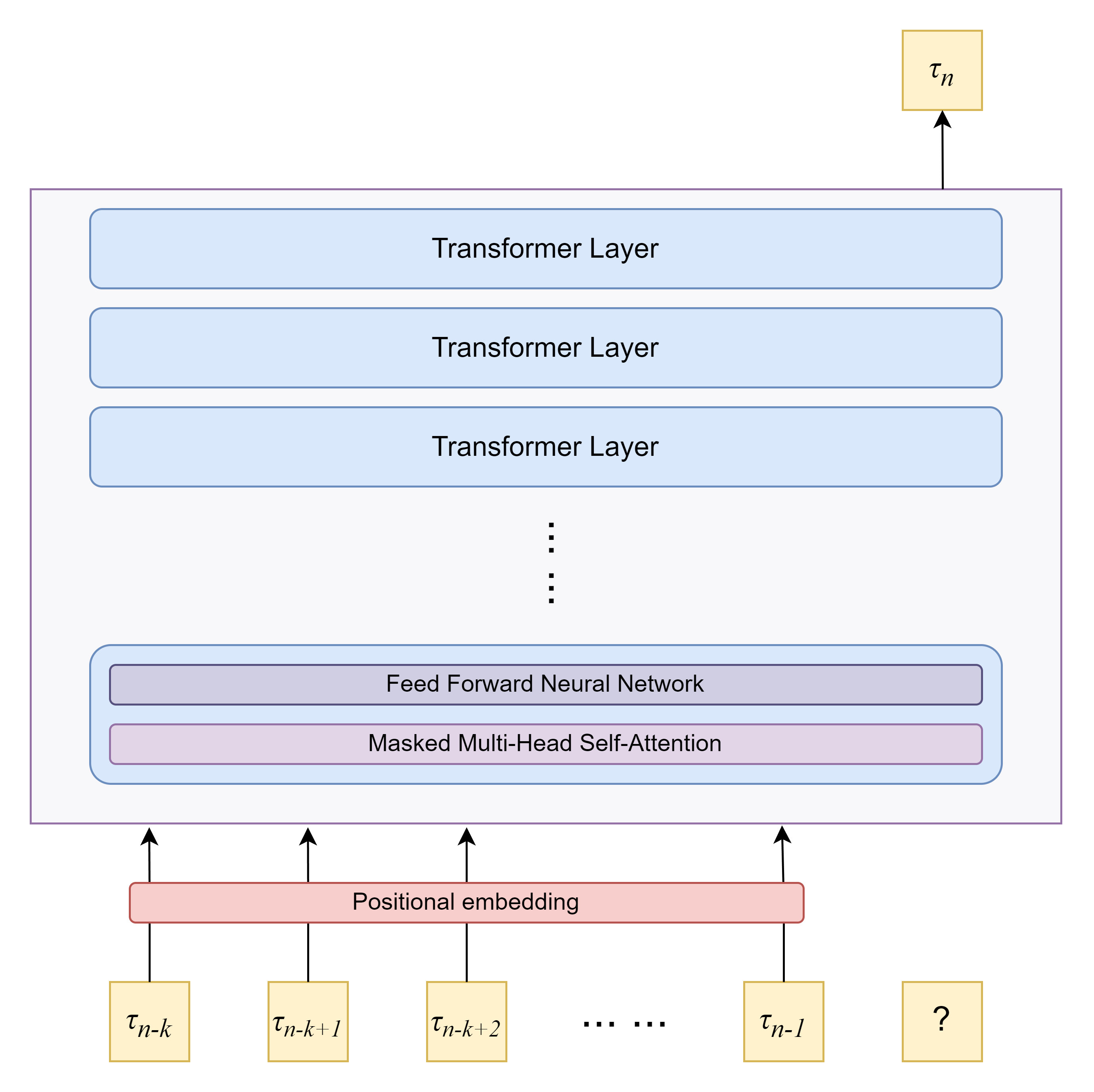}}
\subfigure[Classification as Downstream task]{\label{fig:clf}\includegraphics[width=60mm]{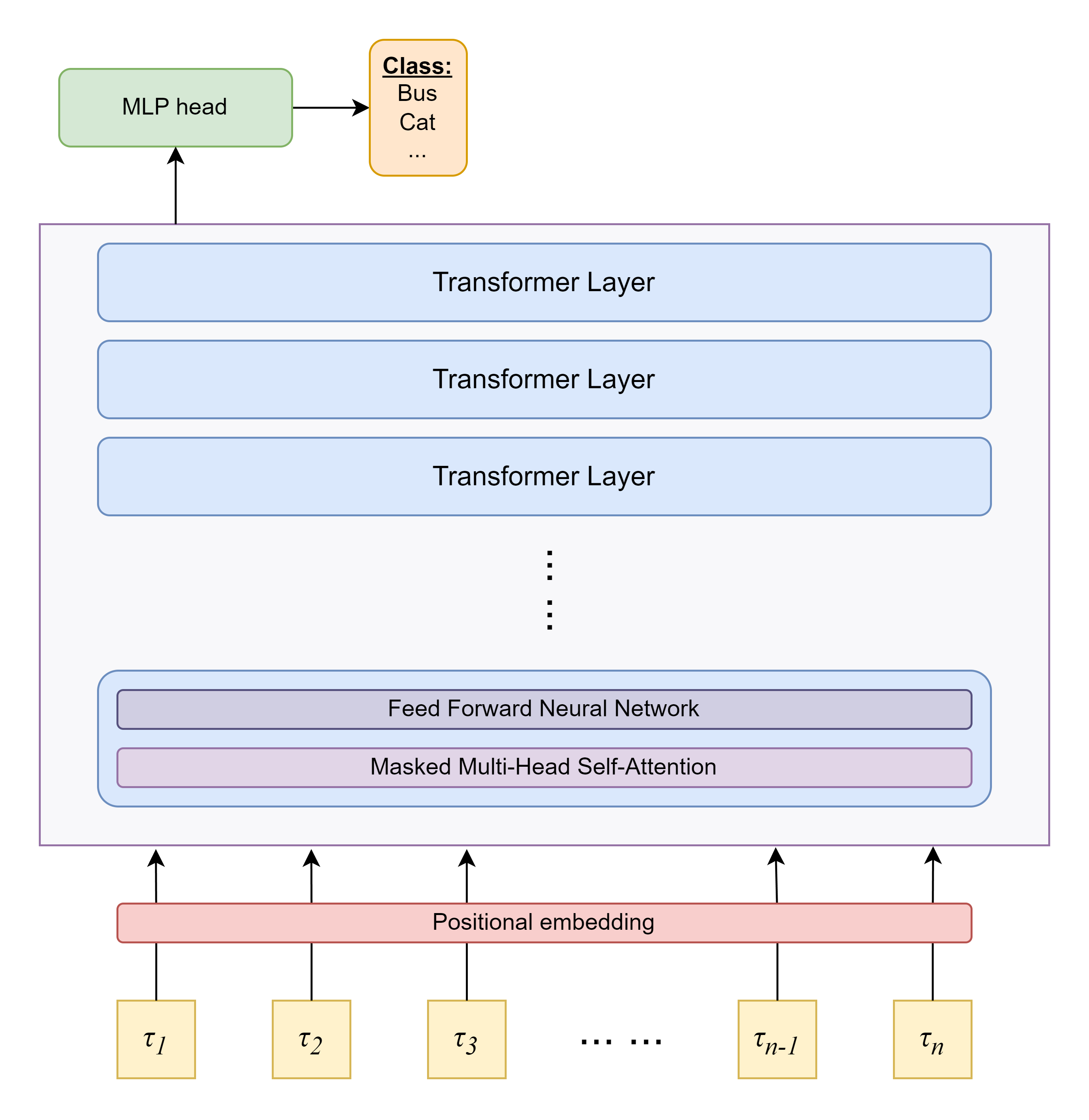}}
\caption{\textbf{Adaptability of SketchGPT.} We illustrate how the sketch completion pre-training phase could later be adapted towards its application for sketch recognition downstream task.}
\end{figure}

Equivalently to the original GPT work \cite{radford2018improving} we use an unsupervised pre-training approach. In this step, our model is exposed to a large corpus of sketch data, and it learns to predict the next token in the stroke sequence using a next token prediction strategy (Fig. \ref{fig:ft}). Given an unsupervised corpus of stroke tokens \(S\), the model predicts the probability distribution of the next token \(\tau_{n}\) given a finite context window of preceding tokens. Thus, the model is trained to minimize the negative log-likelihood loss to predict the correct next token given the previous ones. Formally, the objective function that is minimized is defined as:

\begin{equation}
L_{\text{pretrain}(S)} = \sum_{n=i} -\log P(\tau_{n} | \tau_{n-k}, \ldots, \tau_{n-1} ; \Theta)
\end{equation}

where \(k\) represents the size of the context window, and \(P\) is the conditional probability modeled in the network with parameters \(\Theta\).  This process equips the model with an innate understanding of sequential dependencies, structural nuances, and contextual relationships within sketch data.

\subsection{Learning tasks through fine-tuning}
Once the model has been pre-trained, we explore the ability of this model to be utilized for downstream tasks, through a fine-tuning process. In particular we focus in the following tasks:

\begin{itemize}
\item \textbf{Sketch completion and generation:} In this first task, the model is trained to complete and generate sketches autoregressively. The fine-tuning process consists of feeding the model with partially drawn sketches and training it to sequentially complete the missing components through a next token prediction framework within a specific class context (unlike the pre-training step where the model is trained across the entire corpus without class discrimination). This task inherently encompasses unconditional sketch generation too, as a special case, where the model learns to generate a sketch autoregressively from scratch without any explicit input.

\item \textbf{Sketch classification:} This task, illustrated in Fig. \ref{fig:clf}, takes as input a sketch \(S_i\) and predicting it's category label \(y_i\). We assume a dataset \(D\), where for each instance of that dataset we have the sequence of input tokens \( \tau_1, \ldots, \tau_n\) and its label \(y\).  To fine-tune the model for the task, we pass the inputs through our pre-trained model to obtain the final GPT block’s activation, which is then fed into the classification head to predict \(y\). This leads to the following objective function to minimize in the fine-tuning process:
\begin{equation}
L_{\text{classification}(D)} = \sum_{(S, y)} -\log P(y | \tau_1, \ldots, \tau_n)
\end{equation}

\end{itemize}

\section{Experimental Validation}\label{s:results}



We extensively tested our SketchGPT model to assess its performance in the proposed downstream tasks. Our goal is to demonstrate the feasibility of using autoregressive models for sketch-related tasks, employing simplified representations through sketch abstraction with primitives. We conduct various experiments to highlight the effectiveness of our approach, analyzing results both quantitatively and qualitatively. 

\subsection{Datasets}
The proposed model is evaluated on the QuickDraw dataset~\cite{zhong2019publaynet}, which consists of over 50 million hand-drawn sketches across 345 different categories in stroke-3 format. Collected from the Quick, Draw! game by Google, it comprises sketches drawn by humans within a 20-second limit. The data in available in two formats: images and stroke-3. For our work we make use of the latter.






\subsection{Sketch Generation and Completion}

In this initial set of experiments, we focus on evaluating the SketchGPT's capacity for sketch generation and completion. \medskip 


\noindent
\textbf{Competitor.} To evaluate the generation capacity of our model we compare it to the SketchRNN, a RNN-based seq-to-seq variational autoencoder for vector drawings. To perform the evaluation a subset form the \textit{QuickDraw} dataset is used. Specifically, we chose 7 classes based on the availability of pre-trained weights provided by the SketchRNN authors. The 7 classes are: bus, cat, elephant, flamingo, owl, pig and sheep.

\noindent
\textbf{Evaluation method.} We make use of a CNN-based approach, employing a multi-class ResNet34 trained to classify images of the 7 selected classes from the QuickDraw dataset. The idea is to rasterize the generated sketches from SketchRNN and our method into images, and feed them to the trained CNN. This way we can discern which method produces sketches that are more readily recognizable, as reflected by the CNN's top-1 accuracy (following the common practice in \cite{DBLP:conf/cvpr/SongPSXH18} and \cite{DBLP:conf/bmvc/SuQPYS20}) and top-3 accuracy. More concretely, we generate 1000 sketches for each of the 7 classes using both models, resulting in a total of 7,000 generated samples for each model. ResNet34 classifier is then used to  assess the quality of each approach's generation. ResNet34 was trained using an early stopping strategy to prevent overfitting, with the model reaching a top-1 validation accuracy of 87.92\% at the stopping point.

\noindent
\textbf{Results and Discussion.}
Table \ref{tab:cnn_results_summary} shows the performance results for the two compared models in terms of the selected metrics. Notably, SketchGPT achieves a higher performance for both top-1 and top-3 accuracy, indicating a more finer generation ability for the selected classes. Interestingly, given that the CNN used to asses the models was trained on sketches created by humans (using the QuickDraw dataset), intuitively, this comparison also serves as an indicator of which method yields sketches more similar to those drawn by humans. In this case, the performance of the two models suggests that SketchGPT slightly improves on SketchRNN generating more recognizable and human-like sketches. \smallskip

\begin{table}[t]
\centering
\caption{\textbf{Quantitative comparison} between SketchRNN and SketchGPT recognition accuracy using a CNN classifier.}
\begin{tabular}{lcc}
\toprule
Method                                      & Top-1 acc.                        & Top-3 acc.            \\ \midrule
SketchRNN & 44.6\% & 79.1\% \\ 
SketchGPT & \textbf{50.4\%} & \textbf{81.7\%} \\ \bottomrule
\end{tabular}
\label{tab:cnn_results_summary}
\end{table}

\begin{figure}[t]
    \centering
    \includegraphics[width=0.7\linewidth]{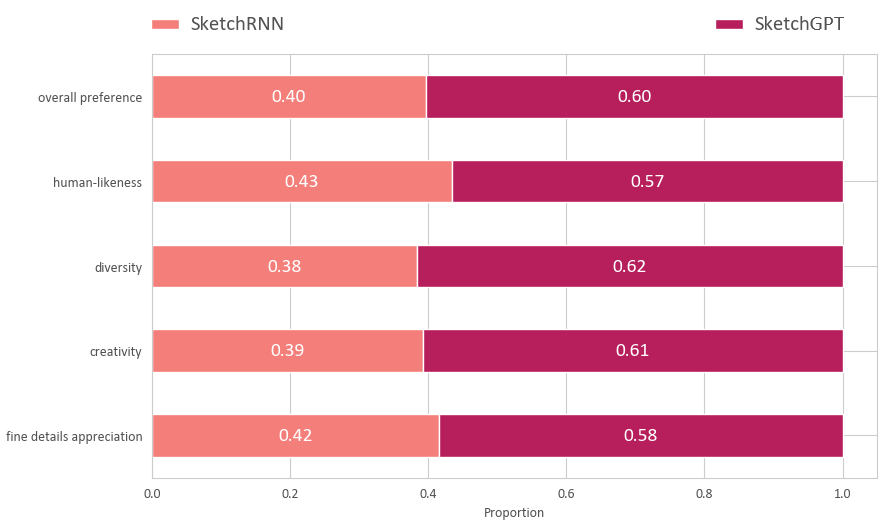}
    \caption{\textbf{Human User Study} results conducted based on the five properties shown in the legend.}
    \label{fig:user_study}
\end{figure}

\subsection{Human Evaluation Study}

This experiment aims to evaluate the variability, human plausibility and overall aesthetic of the sketches generated by our model compared to SketchRNN. The study involved 100 individuals, showing them five sketches generated from scratch using both SketchGPT and SketchRNN, for each of the seven classes of the QuickDraw dataset whose pre-trained weights for SketchRNN are publicly available (bus, cat, elephant, flamingo, owl, pig and sheep). Each participant was given 5 different questions regarding the two compared models to evaluate different qualitative aspects. More concretely, the questions aimed to assess the two models based on the following criteria: (1) fine details appreciation, (2) creativity, (3) diversity, (4) human-likeness and (5) overall preference. As can be observed in Fig. \ref{fig:user_study}, SketchGPT consistently surpasses SketchRNN in all five examined properties. Notably, it excels in aspects of diversity and creativity, showcasing a superior capability to generate a broader and richer spectrum of sketches compared to SketchRNN, which leans towards more stable but less varied outputs.

\begin{figure}[t]
    \centering
    \includegraphics[width=0.9\linewidth]{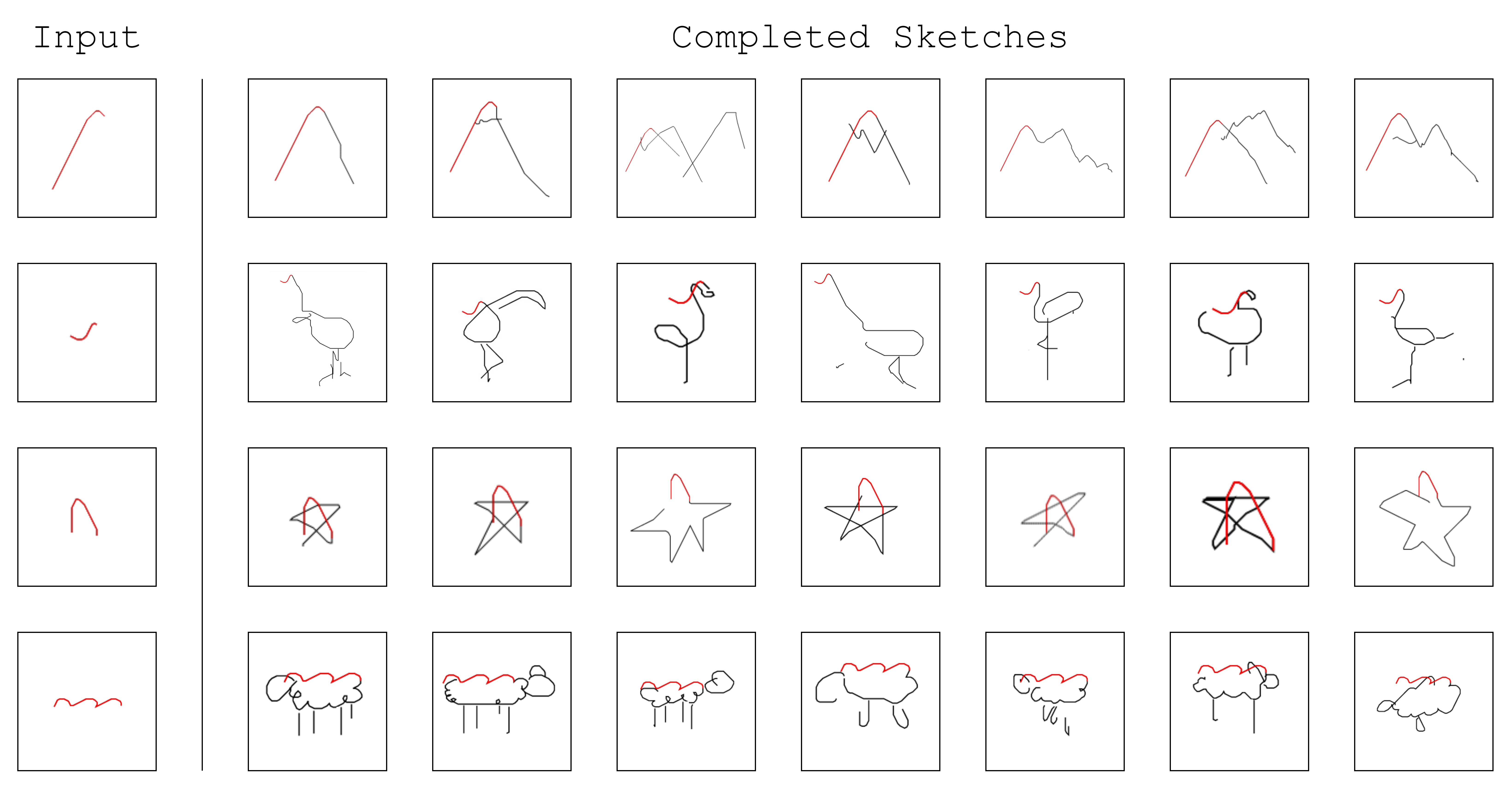}
    \caption{\textbf{Qualitative Analysis for Sketch Generation.} We illustrate the prediction of different possible completion of multiple incomplete sketches.}
    \label{fig:sketch_completion}
\end{figure}

\subsection{Qualitative Analysis for Sketch Completion} 

As discussed earlier, besides generation, SketchGPT also shows promising capacity in sketch completion tasks. In this section, we explore the expressive sketch completion capabilities of our model. By inputting an uncompleted sketch to our model, we generate diverse and plausible completion variations of the given partial sketch. This unique aspect of SketchGPT underscores the model's creativity and adaptability in envisioning different conceivable outcomes. We show results in Fig. \ref{fig:sketch_completion}, using multiple models trained on individual classes. With this experimentation study, we highlight the flexibility and richness of the sketch completion process by our proposed methodology.

\subsection{Sketch Recognition}

\noindent
\textbf{Competitors.} We compare SketchGPT to several existing baselines. (1) HOG-SVM \cite{eitz2012humans}: Classical approach that makes use of HOG features and SVM to perform the classification task. (2) Ensemble \cite{li2013sketch}: Makes use of multiple sketch feature to perform the classification task. (3) Bi-LSTM \cite{hochreiter1997long}: A four-layered bi-directional LSTM with the dimension of the hidden states being 256, used to perform sketch recognition. (4) Sketch-a-Net \cite{yu2016sketch}: A multi-scale multi-channel deep CNN framework designed to perform sketch related tasks. (5) DSSA \cite{song2017deep}: An attention module and a high-order energy triplet loss function are introduced to enhance the original Sketch-A-Net model. (6) ResNet \cite{he2016deep}: A popular CNN architecture highly employed in classification tasks. (7) SketchBERT \cite{lin2020sketch}: A model that generalizes BERT to sketch domain. (8) ViT \cite{dosovitskiy2020image}: A transformer-based neural network architecture tailored for computer vision tasks.

\noindent
\textbf{Evaluation method.} For the evaluation, a subset of classes of the \textit{QuickDraw} dataset is used. We compare the results of training our model using 100 classes, following the evaluation procedure proposed in the SketchBERT paper \cite{lin2020sketch}. We train the SketchGPT for classification task of 100 categories with 5K train samples, 2.5K validation samples and 2.5K test samples to show a fair comparison with the proposed baselines results. The performance is evaluated with Top-1 and Top-5 accuracy metrics. Furthermore, we show the ablation study of our model with and without pre-training, to showcase the effect of the process of pre-training in the performance of the model.

\noindent
\textbf{Results and Discussion.} 
In Table \ref{tab:clf_results}, we can observe the comparative performance of our classification model against several baselines in terms of Top-1 and Top-5 accuracy. The 1st/2nd best results are indicated in \color{red}red\color{black}/\color{blue}blue\color{black}. SketchGPT achieves the second-highest performance for both of the metrics, getting surpassed only by SketchBERT. It showcases superior recognition capabilities compared to conventional classification methodologies, prevalent CNN-based frameworks designed for sketch recognition, and even surpasses state-of-the-art models such as ViT. However, regarding it's comparison with SketchBERT, it is important to note that a fair comparison with the aforementioned baseline is hard to achieve, principally due to the lack of knowledge regarding the specific classes used for evaluation. SketchBERT's paper does not specify the 100 classes utilized, potentially leading to a disparity in complexity between the classes evaluated. Furthermore, SketchBERT's reported results are based on pre-training with 345 classes and 70K samples, followed by fine-tuning for classification with the 100 classes. In contrast, our pre-training involved only 50 classes and 5K samples, a considerably smaller dataset, principally due to computation limits. Notably, SketchBERT's ablation studies indicate that using smaller pre-trained datasets can lead to inferior results, as evidenced by their experiments with 100 classes and 5K samples resulting in 85.82\% top-1 accuracy and 97.31\% top-5 accuracy. Given all these factors, it is reasonable to assert that a more equitable comparison, utilizing the same classes and employing larger pre-training datasets for our model, may yield even more favorable results for SketchGPT.

\begin{table}[htp]
\centering
\caption{\textbf{Top-1 and Top-5 accuracy for Sketch Recogntion} when compared to other baselines; w./o., and w indicate the results without and with the pre-training process.}
\medskip
\begin{tabular}{
    l                   
    S[table-format=1.6] 
    S[table-format=1.5] 
}
\toprule
\textbf{Methods}
& {\qquad\qquad\qquad\qquad\qquad \textbf{QuickDraw (\%)}} \\
\midrule
 &  
{\textbf{Top-1 acc.}} & 
{\textbf{Top-5 acc.}}\\
\midrule
HOG-SVM \cite{eitz2012humans} & 52.05 & 74.50\\
\midrule
Ensemble \cite{li2013sketch} & 60.31 & 80.22\\
\midrule
Bi-LSTM \cite{hochreiter1997long} & 74.68 & 90.59\\
\midrule
Sketch-a-Net \cite{yu2016sketch} & 70.64 & 87.93\\
\midrule
DSSA \cite{song2017deep} & 79.47 & 92.41\\
\midrule
ResNet18 \cite{he2016deep} & 79.67 & 91.71\\
\midrule
ResNet34 \cite{he2016deep} & 82.02 & 93.48\\
\midrule
SketchBERT \cite{lin2020sketch} & \color{red}88.30 & \color{red}97.82\\
\midrule
ViT \cite{dosovitskiy2020image} & 47.69 & 68.73\\
\midrule
\midrule
SketchGPT (w/o PT) & 81.42 & 91.81\\
\midrule
SketchGPT (w PT) & \color{blue}83.58 & \color{blue}93.65\\
\bottomrule
\end{tabular}
\label{tab:clf_results}
\end{table}

\subsection{Ablation studies}

In this section, we conduct multiple ablation studies to analyze how our models performance is affected by altering different conditions.

\begin{figure}[h]
    \centering
    \includegraphics[width=\linewidth]{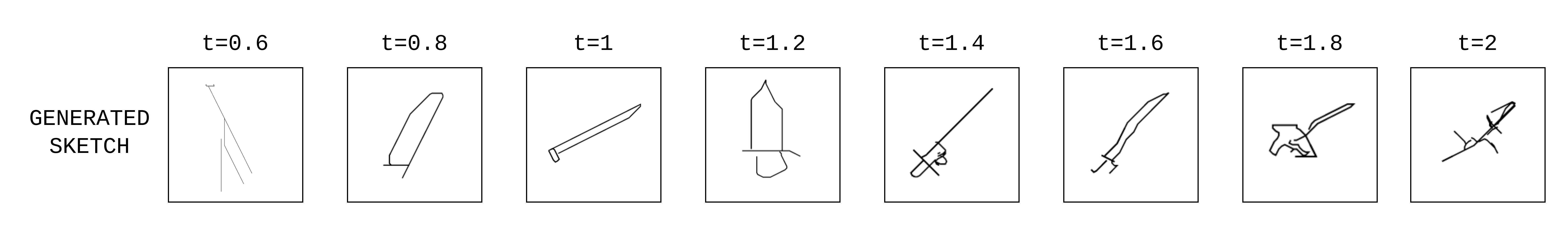}
    \caption{\textbf{Temperature Scaling} impact on our model's generated sketches.}
    \label{fig:temp_study}
\end{figure}

\noindent
\textbf{Model's generation ability through temperature (\textit{t}) modulation.} In this experiment, we systematically vary the temperature parameter of our model for the generation task from 0.6 to 2.0. The results are shown in Fig. \ref{fig:temp_study} for the \textit{sword} class. The temperature parameter controls the degree of randomness in sampling from the model's output probability distribution. Upon analyzing the results, we notice that at lower temperatures (from 0.6 to 0.8), the model produces abstract and too simplistic sketches with minimal detail, while at higher temperatures (1.6 to 2.0), it tends to generate more random and nonsensical compositions.
Interestingly, within the range of 1.0 to 1.4, the model strikes a balance, generating sketches with greater coherence and complexity. This showcases how temperature plays a major role in influencing the diversity and quality of generated outputs.

\noindent
\textbf{Impact of total class counts on classification performance.} In the context of the classification task, this experiment analyzes the change in performance of our model as we increase the number of classes used to train the model. The study is done training our models with the same settings, but just changing the class counts from 25 to 200. For each number of categories, we use 5K samples per class to train the model. In Table \ref{tab:ablation_table} we can observe the results. A slight decrease in performance is observed as the number of categories increases. Nevertheless, this decline is less pronounced from 25 to 50 to 100 classes, with a more noticeable decrease in performance upon reaching 200 classes. Notably, these observations were made under fixed training size conditions, augmenting the training dataset size might offer a viable solution to mitigate the performance disparities observed with higher class counts.

\begin{table}[t]
\tiny\setlength{\tabcolsep}{1pt}
\begin{minipage}{.3\linewidth}
\centering

\caption{\textbf{Sketch Recognition performance} with different class counts/sample sizes/network structures.}
\label{tab:ablation_table}

\medskip

\begin{tabular}{ccccc}
    \toprule
    {\bf Class counts.} & {\bf Top-1 acc.} & {\bf Top-5 acc.} \\
    \midrule
    \textbf{25 classes} & $\textbf{84.89\%}$ & $\textbf{94.21\%}$\\
    \textbf{50 classes} & $83.68\%$ & $94.12\%$\\
    \textbf{100 classes} & $83.58\%$ & $93.65\%$\\
    \textbf{200 classes} & $79.86\%$ & $90.06\%$\\
    \bottomrule
\end{tabular}
\end{minipage}\hfill
\begin{minipage}{.3\linewidth}
\centering

\label{tab:second_table}

\medskip

\begin{tabular}{ccccc}
    \toprule
    {\bf Training size.} & {\bf Top-1 acc.} & {\bf Top-5 acc.} \\
    \midrule
    \textbf{5K} & $84.89\%$ & $94.21\%$\\
    \textbf{10K} & $85.67\%$ & $95.34\%$\\
    \textbf{20K} & $85.69\%$ & $95.68\%$\\
    \textbf{75K} & $\textbf{89.95\%}$ & $\textbf{97.51\%}$\\
    \bottomrule
\end{tabular}    
\end{minipage}\hfill
\begin{minipage}{.3\linewidth}
\centering

\label{tab:third_table}

\medskip

\begin{tabular}{ccccc}
    \toprule
    {\bf Network size.} & {\bf Top-1 acc.} & {\bf Top-5 acc.} \\
    \midrule
    \textbf{4-8-256} & $80.75\%$ & $91.56\%$\\
    \textbf{12-8-256} & $83.65\%$ & $93.32\%$\\
    \textbf{12-12-768} & $84.47\%$ & $94.16\%$\\
    \textbf{8-8-512} & $\textbf{84.89\%}$ & $\textbf{94.21\%}$\\
    \bottomrule
\end{tabular}    
\end{minipage} 
\end{table}

\noindent
\textbf{Influence of training data size on classification performance.} In this study, we aim to explore the effect of increasing the training sample size on our model's performance. Given the practical constraints of the large number of classes, the study is performed in a reduced group of 25 categories, to understand if augmenting the training data within the subset actually affects model performance. The evaluation metrics used are Top-1 and Top-5 accuracy, and the results are shown in Table \ref{tab:ablation_table}. We observe that the model benefits from the increase in the number of samples, yielding a percentage increase of 5.96\% in our Top-1 accuracy when expanding the training samples from 5K to 75K.  Extrapolating from these findings, we anticipate that scaling up the training size would likely result in improved performance with a greater number of classes (e.g. 200 classes, 345 classes, etc.), thereby underscoring the potential benefits of increased sample sizes in enhancing model performance.

\begin{figure}[t]
    \centering
    \includegraphics[width=0.5\linewidth]{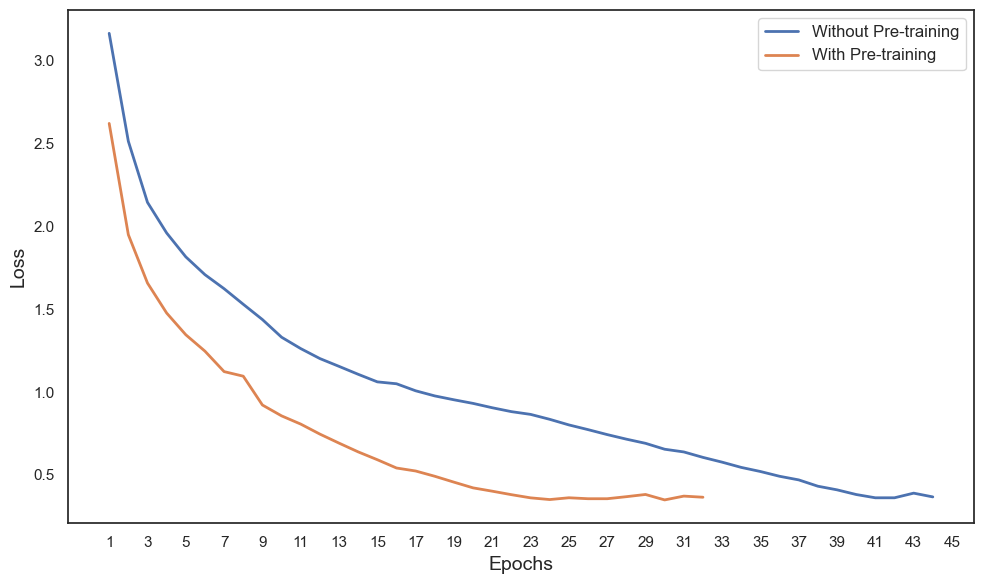}
    \caption{\textbf{Convergence rate of SketchGPT} with and without pre-training.}
    \label{fig:conv_rate}
\end{figure}

\noindent
\textbf{Effects of network's size on classification performance.} We aim to examine how the variation in the size of our model might impact its performance for the sketch recognition task. The parameters explored are the number of layers (L), self-attention heads (A), and the hidden size (H); and four scenarios are investigated to understand the influence. As we observe in Table \ref{tab:ablation_table}, the optimal performance is achieved when L = 8, A = 8 and H = 512, making a good balance between the model complexity and final performance. It is noteworthy that augmenting the number of layers yields a substantive performance boost, elevating top-1 accuracy from 80.75\% to 83.65\%, even with a smaller hidden size of 256. On the other hand, opting for a larger network configuration (L = 12, A = 8, and H = 768) approaches optimal performance too, but the training process duration significantly extends.

\noindent
\textbf{Convergence Rate acceleration through the use of pre-training.}

Parallel to \cite{lin2020sketch} our model also exhibits a faster convergence rate with pre-training on the \textit{QuickDraw} dataset. However, it's worth noticing that the observed acceleration in convergence is way less pronounced in our case. As observed in Fig. \ref{fig:conv_rate}, which shows the loss curves of our model trained on 100 classes with 5K samples per category, after pre-training the model manages to converge from about 40 to 22 epochs. This phenomenon further highlights the importance of pre-training for SketchGPT, demonstrating that besides the boost in performance, we also achieve a less time-consuming training procedure.

\section{Conclusion and Future Work}\label{s:conclusion}
In this work, we have presented a novel approach to employ autoregressive models in the sketch domain. Through an abstraction process that simplifies the sketch data, we facilitate the learning process of our model; demonstrating remarkable proficiency in sketch generation and completion, as well as strong competitiveness in classification tasks. An inherent limitation of our model lies in the data representation, where the abstraction process introduces significant information loss. Future research efforts should prioritize the development of an improved data representation approach to alleviate this challenge. By doing so, we anticipate significant enhancements in model performance and its adaptability to diverse and complex datasets.\\
\noindent
\textbf{Future Scopes and Challenges} Despite the strengths demonstrated by our model in sketch generation and completion tasks, as well as competitive performance in sketch classification, certain limitations are worth noting from our data representation methodology. Although the simplification of data through sketch approximation with primitives aids learning, it also results in information loss which adversely affects performance. Moreover, the absence of techniques like the Ramer-Douglas-Peucker (RDP) algorithm \cite{Ha2018},\cite{lin2020sketch} to shorten the sequence length of the sketches hinders the model convergence, affecting the model performance. Since our approximation method already results in some information loss, we opted against using additional algorithms like RDP, where further simplification could lead to more information loss. Our methodology yields competitive performances when applied to the QuickDraw dataset, characterized by relatively simpler sketches. However, in more complex datasets such as TU-Berlin, \cite{eitz2012humans}, the data approximation strategy through primitives results in some information loss that impacts the model's performance, particularly in tasks like sketch completion and generation. The road forward calls for a more effective data representation strategy, which should minimize information loss while decreasing the maximum sequence length to enhance both model convergence and effectiveness in performance.

\section*{Acknowledgment}
This work has been partially supported by the Spanish project PID2021-126808OB-I00, the Catalan project 2021 SGR 01559, the CVC Internship Program and PhD Scholarship from AGAUR 2023 FI-3-00223. The Computer Vision Center is part of the CERCA Program/Generalitat de Catalunya.

\bibliographystyle{splncs04}
\bibliography{main}
%




\end{document}